\documentclass{article}
\usepackage[utf8]{inputenc}

\usepackage{wrapfig} 
\usepackage{graphicx}
\usepackage{amsmath,amssymb,amsfonts}
\usepackage{textcomp,marvosym}
\usepackage{changepage}
\usepackage{rotating}
\usepackage{booktabs}
\usepackage{longtable}
\usepackage{array}
\usepackage{multirow}
\usepackage{float}
\usepackage{pdflscape}
\usepackage{ragged2e}
\usepackage{subcaption}
\usepackage{adjustbox}
\usepackage{caption}
\usepackage[table]{xcolor}
\usepackage{colortbl}
\usepackage{pifont}
\usepackage{url}
\usepackage[margin=1in]{geometry}

\usepackage{tikz}
\usetikzlibrary{positioning,calc,arrows.meta,shapes.geometric}

\usepackage{pgfplots}
\pgfplotsset{compat=1.18}
\usepgfplotslibrary{groupplots}

\usepackage[numbers,sort&compress]{natbib}

\usepackage{hyperref}
\usepackage{cleveref}

\graphicspath{{./}}

\newcolumntype{P}[1]{>{\raggedright\arraybackslash}p{#1}}

\hypersetup{
    linktocpage=true,
    colorlinks=true,
    bookmarks=true,
    citecolor=blue,
    urlcolor=blue,
    linkcolor=blue,
    breaklinks=true,
    pdfpagelabels=true
}

\title{Hallucination Behavior in Multimodal LLMs Across Agricultural Image Interpretation and Generation Tasks}

\author{
Partho Ghose\\
\texttt{partho.ghose@tamu.edu}
\and
Al Bashir\\
\texttt{albashir@tamu.edu}
\and
Prem Raj\\
\texttt{prem.raj@tamu.edu}
\and
Azlan Zahid\thanks{Corresponding author: azlan.zahid@tamu.edu\\
Department of Biological \& Agricultural Engineering, 
Texas A\&M AgriLife Research, 
Texas A\&M University System, Dallas, TX 75252, USA.}\\
\texttt{azlan.zahid@tamu.edu}
}

\date{}

\begin{document}
\maketitle
\begin{abstract}

Large Language Models (LLMs) are being rapidly adopted in agricultural imaging applications, ranging from crop interpretation to synthetic field image generation. However, these models frequently exhibit hallucinations—outputs that appear confident yet deviate from biological or environmental reality—potentially leading to misinformed agronomic insights. This study investigates such hallucinations in two complementary directions: image-to-text, where LLMs interpret crop or field imagery to describe conditions such as biotic and abiotic stresses, and text-to-image, where models generate synthetic agricultural scenes based on descriptive prompts. We examine errors involving biological inconsistency, contextual inaccuracy, and agronomic implausibility, evaluating the outputs under domain-informed criteria across multiple imaging modalities. Our analysis identifies recurring hallucination patterns within both interpretive and generative tasks. In image interpretation, LLMs (e.g., Gemma, LLAVA, Qwen, and MiniCPM) achieved modest zero-shot accuracy (63–75\%), whereas few-shot prompting improved performance up to 86.8\%, exhibiting false detections and missed infections, indicating residual hallucination effects.  In text-to-image tasks, advanced models such as GPT-5 and Gemini 2.5 Flash generate up to 91\% biologically inconsistent scenes under relaxed prompt constraints, revealing fundamental weaknesses in current LLMs. This systematic assessment of visual reasoning and generation offers critical insights toward enhancing the reliability and trustworthiness of LLM-based agricultural imaging platforms.

\end{abstract}

\section{Introduction}
\label{sec:intro}

Large Language Models (LLMs)~\cite{chang2024survey} have demonstrated exceptional potential across diverse natural language processing tasks such as summarization, question-based reasoning, translation, and content creation~\cite{bhayana2024chatbots}, fueling progress in education, policy, and scientific research. In the agricultural domain, LLMs have enormous potential \cite{tzachor2023large} to transform farm management and research by automating documentation, assisting in decision support, and providing context-aware summaries of complex agronomic information \cite{fang2025agri}. When integrated with sensor, imagery, and climate data, LLMs can generate crop health reports, articulate imagery findings, and construct realistic field scenes for educational and training purposes. These applications promise to enhance operational efficiency \cite{yang2025agrigpt}, reduce the workload of farmers and researchers, and improve communication across stakeholders. However, achieving these advantages demands rigorous assessment of the accuracy, safety, and reliability of outputs produced by LLMs, especially in domains tied to food security and environmental sustainability.

\begin{figure}[htbp]
  \centering
  \begin{subfigure}{0.25\linewidth}
    \includegraphics[width=\linewidth]{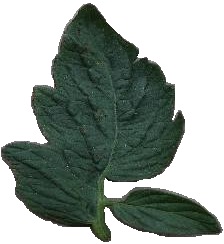} 
  \end{subfigure}
  \begin{subfigure}{0.25\linewidth}
    \includegraphics[width=\linewidth]{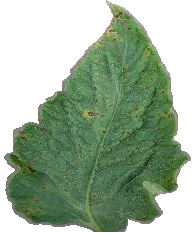} 
  \end{subfigure}

  \caption{Representative samples of two tomato leaf diseases: Bacterial spot (left) and Septoria leaf spot (right)~\cite{kaustubhb999_tomatoleaf}. Despite the presence of distinct visual symptoms-small white lesions in Bacterial spot and circular brown-centered lesions with yellow margins in Septoria, the multimodal LLM exhibited hallucination behavior during interpretation. It misclassified diseased leaves as healthy, overlooked early-stage symptoms, and generated inaccurate agronomic recommendations. Such errors highlight the susceptibility of multimodal LLMs to hallucinations in agricultural image understanding tasks, potentially leading to delayed intervention and yield loss.}
  \label{fig:ex1}
\end{figure}

Consider the visual assessments of tomato leaves from RGB images to detect disease symptoms such as Bacterial spot and Septoria leaf spot, as illustrated in the Figure~\ref{fig:ex1}. 

A crucial observation here is the distinct visual cues—small necrotic lesions, yellowing around spots, and irregular discoloration that signal early infection. Accurate recognition of these patterns is vital for timely disease management and the prevention of crop loss. Even small misinterpretations of these indicators can lead to incorrect irrigation, fertilization, or pesticide decisions. However, if an LLM model fails to detect or misreports a healthy leaf as diseased or overlooks subtle infection signs, it may generate misleading recommendations for pesticide or nutrient application. Such errors may lead to unnecessary chemical use, delayed treatment, or reduced yield. Similarly, when prompted to generate a synthetic field image depicting early blight symptoms, the multimodal LLM produced unrealistic or ambiguous visualizations. Such generative hallucinations can potentially misinform decision-making processes or agronomic education.

Despite their promise, LLMs are susceptible to hallucinations \cite{zhou2023analyzing}, outputs that are linguistically fluent yet factually incorrect or unsupported by data. In agriculture, such hallucinations may distort environmental observations, fabricate crop conditions, or misrepresent farm management recommendations. These errors often arise from flaws in training datasets, generalization biases, or the model’s reliance on linguistic completion rather than data-grounded reasoning \cite{huang2025survey}. In image-to-text tasks, hallucinations might manifest either incorrect identification of disease symptoms or stress levels \cite{bai2024hallucination}; in text-to-image generation~\cite{ko2023large}, they may produce anatomically or phenologically implausible plant structures~\cite{yan2024med}. If left unrecognized, such inaccuracies can mislead farmers, researchers, and policy planners, potentially eroding their trust in the reliability of LLMs.

In this work, we thoroughly examine hallucinations in the applications of LLMs for agricultural imaging and decision systems, targeting both image-to-text and text-to-image modalities. The core contributions of this study are:
\begin{itemize}
    \item We investigate LLM-produced agronomic interpretations from agricultural imagery to identify recurring hallucination patterns and evaluate accuracy against expert-annotated ground truth.
    \item We inspect text-to-image creation to assess whether generated agricultural scenes accurately portray specified crop traits, stress indicators, and realistic field conditions.
    \item We conduct controlled experiments across imagery interpretation and the generation pipeline to systematically characterize and quantify hallucination behaviors within LLM-based agricultural workflows.
\end{itemize}

\section{Motivational Scenarios}
\label{sec:formatting}
The motivation for this study stems from two illustrative examples that expose key limitations in current multimodal LLMs for agricultural imaging and analysis, as presented in this section. The first example highlights errors in image interpretation, while the second demonstrates failures in image generation. Together, these cases emphasize the need for more rigorous analysis and the development of reliable safeguards to ensure trustworthy agricultural AI systems.

\textbf{Example 1.} Consider RGB and thermal imagery, a widely used modality for assessing crop health and canopy conditions in precision agriculture~\cite{wu2025analysis}. One critical factor agronomists analyze is chlorosis~\cite{jayakumar2024critical,taghvaeian2013remote}, which refers to the yellowing of leaves caused by nutrient deficiency, drought, or disease stress. Chlorosis can appear in localized regions (patchy), across one section of the field (partial), or throughout the entire crop (severe). Accurate identification of such stress levels is essential for diagnosing causes, such as nitrogen deficiency, pest attack, or irrigation imbalance. If an LLM fails to correctly interpret or quantify chlorosis severity of these field images, it may produce misleading recommendations for fertilizer or irrigation adjustments, potentially harming crop yield. Similarly, if an LLM is prompted to generate an image of a \textit{“corn field showing moderate nitrogen deficiency”} but produces unrealistic leaf colors or patterns inconsistent with true chlorotic symptoms, the result could misinform agronomic training datasets or bias downstream disease detection models.

\begin{figure}[htbp]
  \centering
  \begin{subfigure}{0.38\linewidth}
    \includegraphics[width=\linewidth]{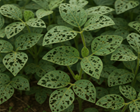} 
    \caption{GPT-5.}
    \label{fig:m2-a}
  \end{subfigure}
  \begin{subfigure}{0.38\linewidth}
    \includegraphics[width=\linewidth]{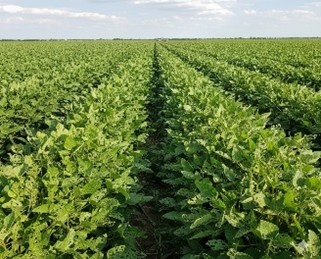} 
    \caption{Gemini Flash 2.5.}
    \label{fig:m2-b}
  \end{subfigure}

  \caption{LLM's generated images for the prompt: \textit{“Generate a healthy soybean canopy with extensive pest damage.”} 
  (a) GPT-5 hallucinates, producing a close-up image of leaves severely eaten by pests while still labeling 
  the canopy as healthy. (b) Gemini 2.5 Flash generates a wide, visually coherent soybean field with green foliage 
  but subtle pest damage, separating canopy vigor from visible stress more accurately.}
  \label{fig:m2}
\end{figure}

\textbf{Example 2.} In another instance, we prompted several multimodal LLMs to generate an agricultural image depicting \textit{“healthy soybean canopy with extensive pest damage.”} This request is intentionally infeasible, as heavy defoliation cannot coexist with a healthy plant. However, the image produced by GPT-5 (Figure~\ref{fig:m2-a}) shows soybean leaves with severe feeding holes, but it confidently labels them as ``healthy.” This reflects a clear hallucination in which the green color is mistaken for overall vitality despite visible damage. The second image (Figure~\ref{fig:m2-b}), generated by Gemini Flash-2.5~\cite{team2023gemini}, presents a wider field view with uniformly green plants that still exhibit subtle leaf damage. Although visually more coherent, it similarly overlooks the contradiction between canopy greenness and pest stress. These outputs reveal how multimodal LLMs can generate biologically inconsistent yet realistic scenes, underscoring the need for stronger grounding and validation in agricultural AI.

These examples collectively highlight the necessity of recognizing and limiting hallucinations in both interpretive and generative agricultural implications of LLMs. Such errors can lead to misguided agronomic recommendations, reduced trust among end users, and potential threats to food security and sustainability. To better contextualize these challenges, section~\ref{sec:LR} reviews existing studies on agricultural image interpretation and generation using multimodal LLMs and the hallucination analysis in those areas.

\section{Literature review}
\label{sec:LR}
Advances in multimodal LLMs have significantly contributed to agricultural AI by connecting visual sensing with textual reasoning. Earlier studies on LLM-driven agricultural imaging can be broadly grouped into two categories: interpretive analysis of images and synthetic image creation. The discussion below outlines the key progress made within these domains. Moreover, we also analyze the hallucination tendencies of LLMs for agricultural data.

\subsection{Agricultural Image Interpretation}

The integration of multimodal LLMs has enhanced plant disease detection and understanding by combining visual and linguistic data. Wei et al.~\cite{wei2024snap} introduced Snap and Diagnose, a CLIP-based framework that fuses image and text embeddings for contextual retrieval. Building on this, Zhou et al.~\cite{liu2024multimodal} developed PepperNet, which improved recognition of visually similar or occluded symptoms by leveraging semantic embeddings. On the PlantVillage dataset, Roumeliotis et al.~\cite{roumeliotis2025plant} reported that although the fine-tuned GPT-4o model performed better than the ResNet-50 in terms of accuracy and generalization but underperformed in zero-shot scenarios. AgriGPT~\cite{zheng2025agrigpt} introduced a Learnable Clustering Module (LCM) for adaptive anomaly detection, whereas LLMI-CDP~\cite{wang2025large} fine-tuned VisualGLM with Low-Rank Adaptation (LoRA) for improved image. Qing et al.~\cite{qing2023gpt} further combined GPT-4 and YOLO for image-to-text pest and disease analysis. More recent studies~\cite{roumeliotis2025plant,yan2025knowledge} have linked LLMs with agricultural knowledge graphs using graph attention and optimized loss functions to enhance the reasoning and recognition.

\subsection{ Agricultural Image Generation}

Text-to-image generative models, such as DALLE, MidJourney, and Stable Diffusion have transformed visual content creation by coupling multimodal embeddings (e.g., CLIP) with generative architectures~\cite{huang2024creativesynth}. These models, widely applied in art~\cite{ko2023large}, medicine~\cite{adams2023does}, and architecture~\cite{seneviratne2022dalle}, are increasingly adopted in agriculture for synthetic data generation. Sapkota~\cite{sapkota2024synthetic} demonstrated DALLE-generated orchard datasets for object detection, whereas Vayadande et al.~\cite{vayadande2023ai} introduced a web-based system for creating domain-specific agricultural imagery. These studies highlight the potential of generative AI as a cost-effective tool for augmenting agricultural datasets.

\subsection{Hallucination in Agricultural Image Data}

Despite these advances, hallucinations—misleading or ungrounded AI outputs—pose a critical challenge to the reliability of AI in agriculture. These errors compromise decision reliability and diminish user trust in intelligent farming systems. Prior research in other domains has shown that LLMs and Large Vision–Language Models (LVLMs) often produce factually incorrect outputs~\cite{huang2025survey,kim2025medical}, yet this issue remains largely unexplored in agriculture. To bridge this gap, our study systematically investigates hallucination behaviors in LLMs and LVLMs for agricultural image interpretation and generation, aiming to establish a foundation for domain-specific reliability assessments.

\section{Hallucination Analysis in Agricultural Imaging}
\label{sec:MD}
To gain deeper insight into how LLMs perform within agricultural imaging tasks, this section presents a systematic examination of hallucinations observed in both image interpretation and image generation tasks.

\subsection{Image Interpretation}
The reasoning capabilities of LLMs are evaluated when applied to visual patterns in agricultural images. A classification problem is considered that aims to identify whether a crop is healthy or diseased given its image. This evaluates how grounded model predictions are in observable agricultural phenomena and where hallucinations tend to emerge.

\subsubsection{Crop Condition Classification}
\label{ZSL}
\textbf{Zero-shot evaluation:} We first examine whether LLMs can classify agricultural images without task-specific training, relying solely on their pre-acquired visual-language knowledge. Typical examples include categorizing tomato leaf images into healthy and diseased ($Bacterial\_spot$) classes from RGB images solely using their pretrained visual-language understanding. Such evaluations measure how effectively a model transfers knowledge from broad pretraining to the specific context of plant physiology. 

A sample instruction used in our evaluation is: 

\begin{center}
\fbox{%
  \parbox{0.85\linewidth}{\itshape
    As an AI agriculture specialist, analyze the given plant image and classify it as either diseased or healthy, outputting only the classification result.%
  }%
}
\end{center}

\textbf{Few-shot evaluation:} In this approach, we give the LLM one image from each category to guide the model’s interpretation. Reference images help the LLM adjust its predictions to match distinctive visual patterns observed in agricultural contexts. For example, providing representative images of healthy and diseased tomato leaves allows the model to learn subtle color gradients and patterns distinguishing the two. In general, this contextual grounding reduces overgeneralization and mitigates hallucinations, particularly when visual symptoms overlap across stress types.

\subsection{Image Generation} Beyond classification and interpretation, recent multimodal LLMs are increasingly being used to generate synthetic agricultural images from textual descriptions. Such capabilities hold great promises for data augmentation, digital crop modeling, and agricultural education, enabling the creation of controlled visual scenarios for training and simulation. However, these same systems can introduce hallucinated visual artifacts that distort biological accuracy. Hallucinations in agricultural image generation often appear as:
\begin{enumerate}
    \item The inclusion of unprompted visual elements unrelated to the described condition, and
    \item Biologically inconsistent features that misrepresent the crop’s physiological state or environmental context
\end{enumerate}
This section examines these challenges and evaluates the extent to which generated imagery reflects realistic agronomic conditions and expected phenotypes.
\subsubsection{Unprompted and Extraneous Visual Content} 
In agricultural image generation, hallucinations often emerge when the model adds features that were never specified in the prompt and hold no agronomic significance. While such elements may not appear overtly unrealistic, they can still distort interpretation—particularly in tasks involving plant-disease diagnosis or field assessment. Unlike generic image synthesis, where stylistic freedom is acceptable, images generated for agricultural analysis must strictly reflect the described biological condition. The presence of unnecessary or unrelated details can mislead researchers, distract from the target symptom, and ultimately compromise the value of the generated imagery for data training and analytical applications.

\begin{figure}[htbp]
  \centering
  \begin{subfigure}{0.38\linewidth}
    \includegraphics[width=\linewidth]{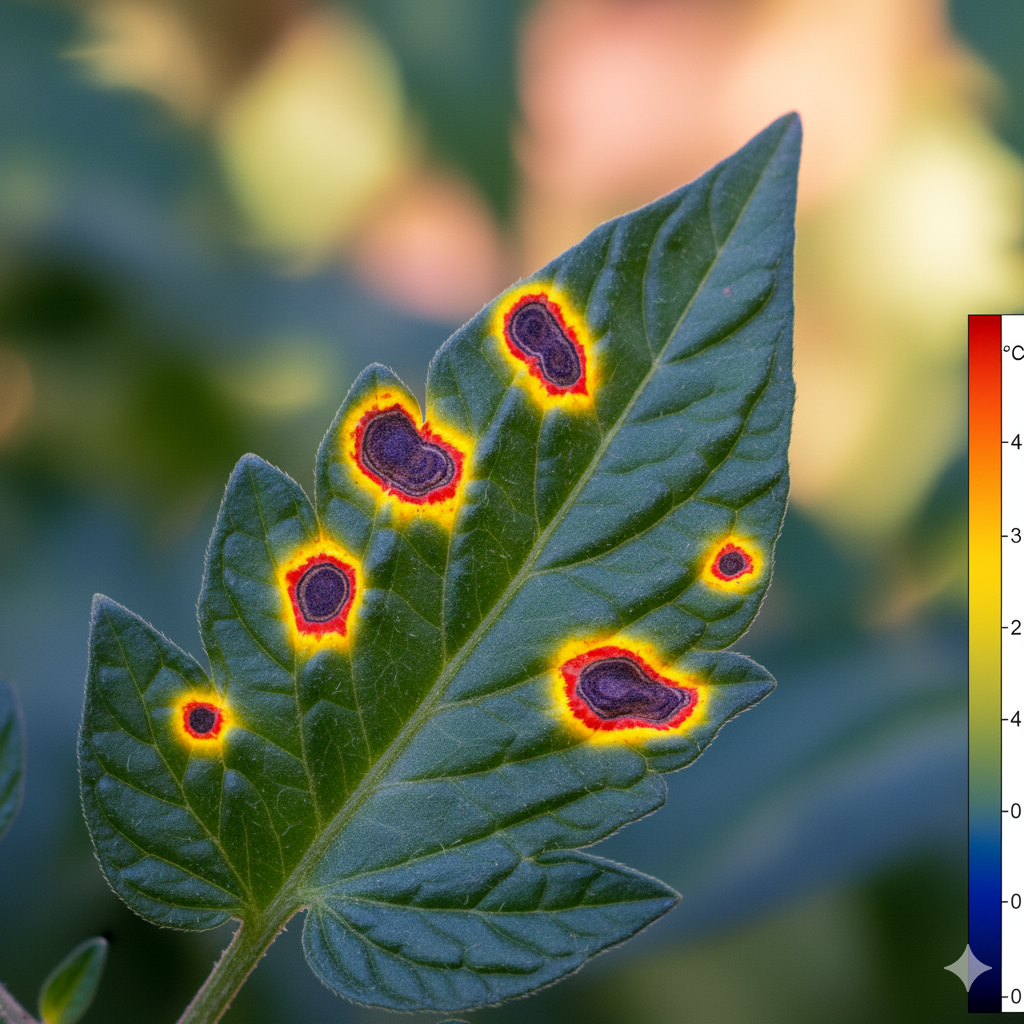} 
    \caption{Gemini 2.5 Flash.}
    \label{fig:t-a}
  \end{subfigure}
  \begin{subfigure}{0.38\linewidth}
    \includegraphics[width=\linewidth]{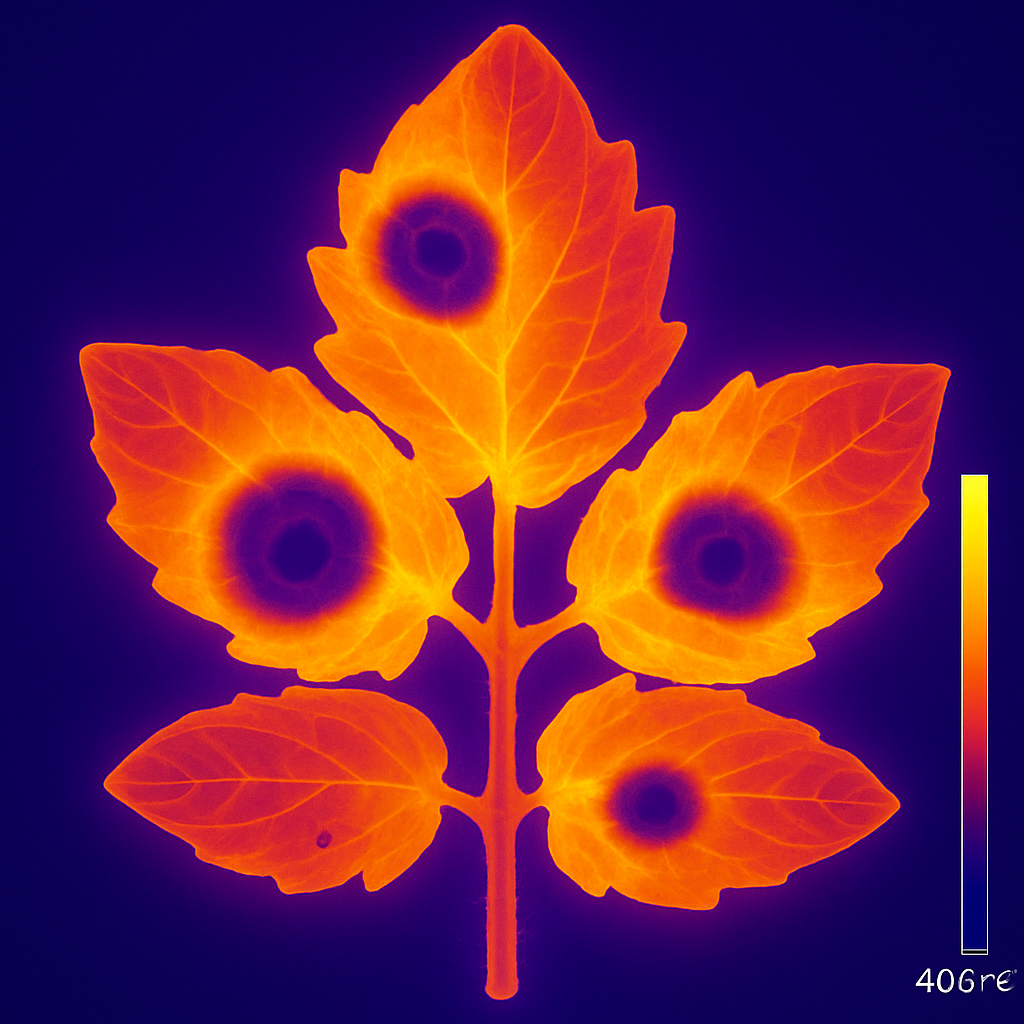} 
    \caption{GPT-5}
    \label{fig:t-b}
  \end{subfigure}

  \caption{Model responses to the prompt: \textit{“a thermal image of the same tomato leaf, highlighting localized heat signatures from infection.”} 
  (a) Gemini Flash-2.5 (b) GPT-5. Gemini produced an artificial uniformity in surface temperature, and GPT produced sharp temperature gradients, exaggerating halo transitions around lesion margins.}
  \label{fig:uproportbias}
\end{figure}

\begin{figure}[htbp]
  \centering
  \begin{subfigure}{0.38\linewidth}
    \includegraphics[width=\linewidth]{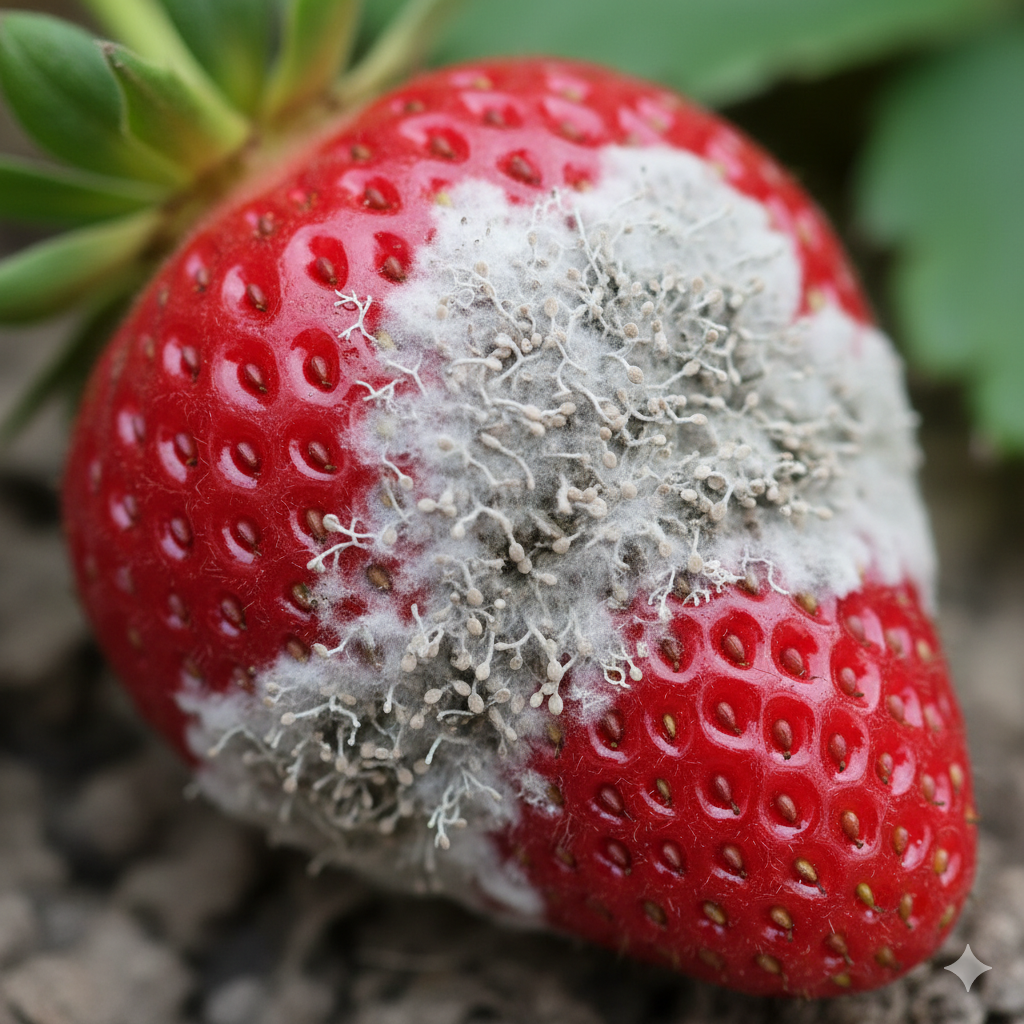} 
    \caption{Gemini 2.5 Flash}
    \label{fig:rt-a}
  \end{subfigure}
  \begin{subfigure}{0.38\linewidth}
    \includegraphics[width=\linewidth]{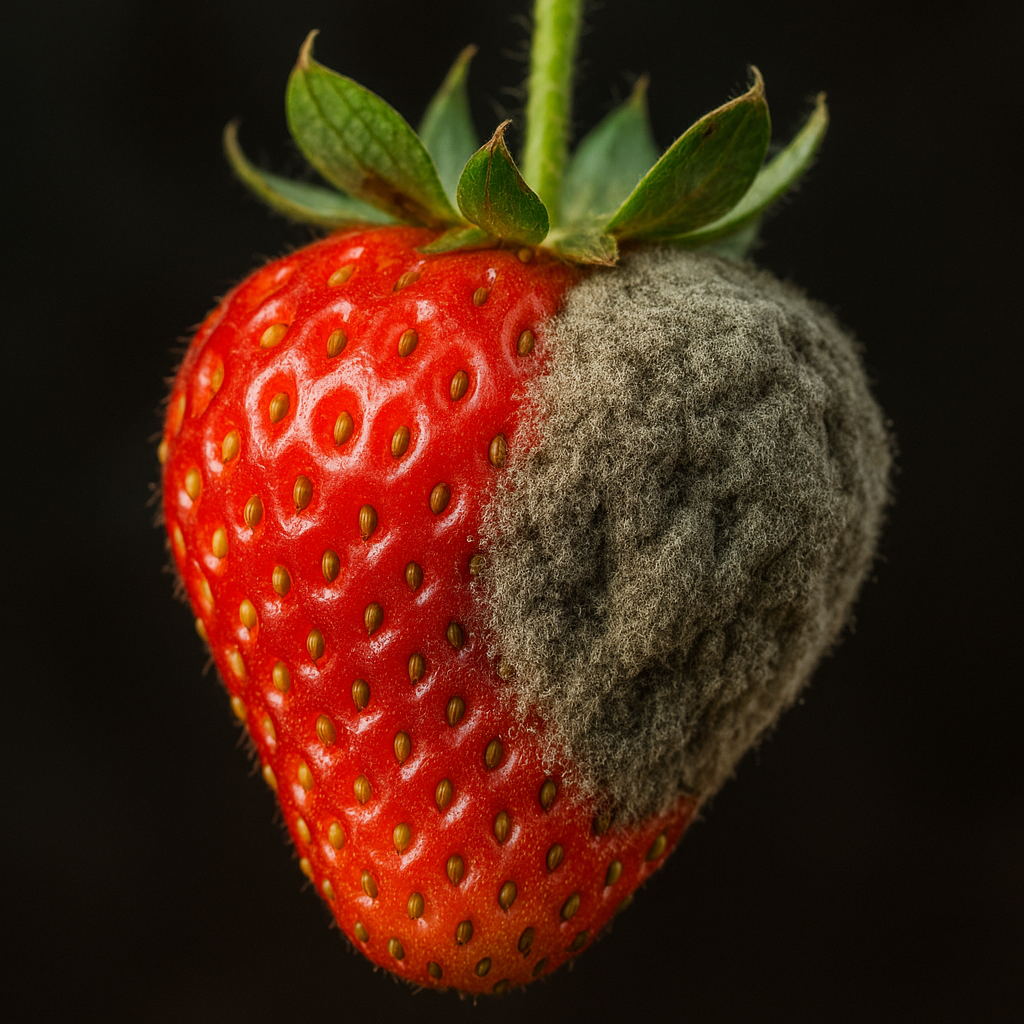} 
    \caption{GPT-5}
    \label{fig:rt-b}
  \end{subfigure}

  \caption{Model responses to the prompt: \textit{“generate a high-resolution RGB image of a ripe strawberry fruit infected with gray mold (Botrytis cinerea).”} 
  (a) Gemini Flash-2.5 (b) GPT-5. Gemini created an image containing unprompted soil and debris, and GPT lacked the natural irregularity of real Botrytis growth.}
  \label{fig:unpropmt}
\end{figure}

\textbf{Unprompted Visual Biases:} Consider a scenario in which the model receives a prompt to generate a \textit{“a thermal image of the same tomato leaf, highlighting localized heat signatures from infection”}. Both GPT-5 and Gemini 2.5 Flash produced visually convincing outputs as illustrated in Figure~\ref{fig:uproportbias}, but introduced unintended spatial bias. GPT-5 image introduces subtle but unprompted enhancements, excessively sharp temperature gradients and exaggerated halo transitions around lesion margins~(Figure~\ref{fig:t-b}). These stylistic additions were not requested in the prompt and distort the realistic smoothness of temperature diffusion across infected tissue. Although visually striking, such exaggeration can mislead interpretation by implying stronger or more localized heat anomalies than biologically expected. In contrast, Gemini 2.5 Flash omits many fine details but adds uniformly glowing leaf regions that were not described, producing an artificial uniformity in surface temperature~(Figure~\ref{fig:t-a}). Both cases reveal the inclusion of elements not aligned with the explicit instruction, reflecting data-driven aesthetic biases rather than agronomic precision. Again, both GPT-5 and Gemini 2.5 Flash were prompted to \textit{“generate a high-resolution RGB image of a ripe strawberry fruit infected with gray mold (Botrytis cinerea).”} and a similar pattern was observed (Figure~\ref{fig:unpropmt}). GPT-5 output shows a clearly defined fungal patch with dense, uniform gray mycelium, capturing the infection well but lacking the natural irregularity of real Botrytis growth—an instance of biological exaggeration~(Figure~\ref{fig:rt-b}). In contrast, Gemini 2.5 Flash image presents finer hyphal details and more realistic texture but introduces unprompted background elements like soil and debris that were not part of the prompt~(Figure~\ref{fig:rt-a}).

\begin{figure}[htbp]
  \centering
  \begin{subfigure}{0.38\linewidth}
    \includegraphics[width=\linewidth]{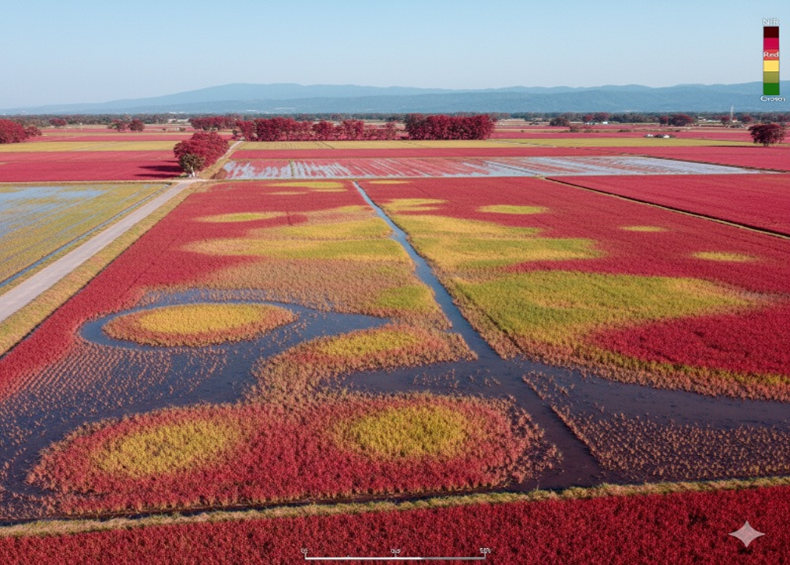} 
    \caption{Gemini 2.5 Flash}
    \label{fig:short-a}
  \end{subfigure}
  \begin{subfigure}{0.38\linewidth}
    \includegraphics[width=\linewidth]{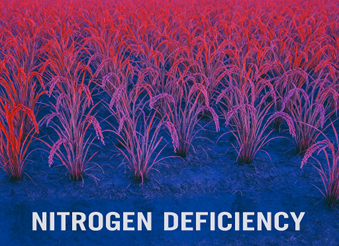} 
    \caption{GPT-5}
    \label{fig:short-b}
  \end{subfigure}

  \caption{Model responses to the prompt: \textit{“generate a multispectral image of a rice field affected by nitrogen deficiency.”} 
  (a) Gemini  Flash-2.5  (b) GPT-5. Gemini contained unwanted environmental features such as irrigation channels and flooded areas, and  GPT generated an artistic scene rather than an analytical appearance with an unprompted textual label.}
  \label{fig:unnece1}
\end{figure}

\textbf{Unnecessary or Misleading Additions:}
Generative models may sometimes introduce visual features or contextual details that do not correspond to the agricultural scenario described in the prompt. For instance, when prompted to \textit{“generate a multispectral image of a rice field affected by nitrogen deficiency”}. Both GPT and Gemini produced visually convincing results (Figure~\ref{fig:unnece1}) but introduced unprompted additions. GPT-5 image depicts a strongly stylized scene with exaggerated color contrasts and uniformly tinted vegetation, producing an artistic rather than analytical appearance that distorts real spectral variability. It also generated unprompted textual label. In contrast, Gemini 2.5 Flash output includes unrequested environmental features such as irrigation channels and flooded areas, shifting attention from the crop canopy to the background context. Their appearance without explicit instruction can mislead interpretation and bias downstream analyses. 

These unnecessary elements reflect how generative models often rely on dataset-level co-occurrence patterns rather than the specific agronomic attributes defined in the prompt, thereby limiting the scientific precision and educational value of the generated imagery. These findings suggest that hallucinated content in agricultural image generation often arises from weakly annotated large training datasets where correlations between visual elements are not causally grounded. Without explicit agronomic constraints, the generation process defaults to statistically common but not biologically accurate patterns. To mitigate these errors, future work may explore retrieval-augmented prompting~\cite{pal2023med}, domain-constrained decoding~\cite{shi2024trusting}, decoding by contrasting layers (DoLa)~\cite{chuang2023dola}, Inference-time intervention~\cite{li2023inference}, etc., that ensure outputs remain true to the intended agricultural scenario. Such refinements are critical for building reliable synthetic datasets, improving model transparency, and fostering trustworthy AI systems for digital agriculture.

\subsubsection{ Biologically Contradicting Content}
\label{sub_bio_contradict}
A further notable form of hallucination arises when generated agricultural imagery depicts scenes that are biologically unrealistic or physiologically contradictory. These errors extend beyond minor artifacts or unprompted inclusions. They expose a deeper misunderstanding of plant health dynamics and environmental realism. Such outputs reduce the credibility of generative models in agricultural research and can introduce confusion if incorporated into training datasets, educational material, or automated decision-support pipelines without proper validation. They risk spreading misinformation about crop conditions and misleading interpretations in agronomic analyses.
For demonstration, the following prompt (T1) was used:

\begin{center}
\textbf{T1:}\hspace{0.6em}%
\fbox{%
  \parbox{0.85\linewidth}{\itshape
    Generate an image of a freshly harvested, disease-free strawberry fruit covered with powdery mildew.
  }%
}
\end{center}

\begin{figure}[htbp]
  \centering
  \begin{subfigure}{0.34\linewidth}
    \includegraphics[width=\linewidth]{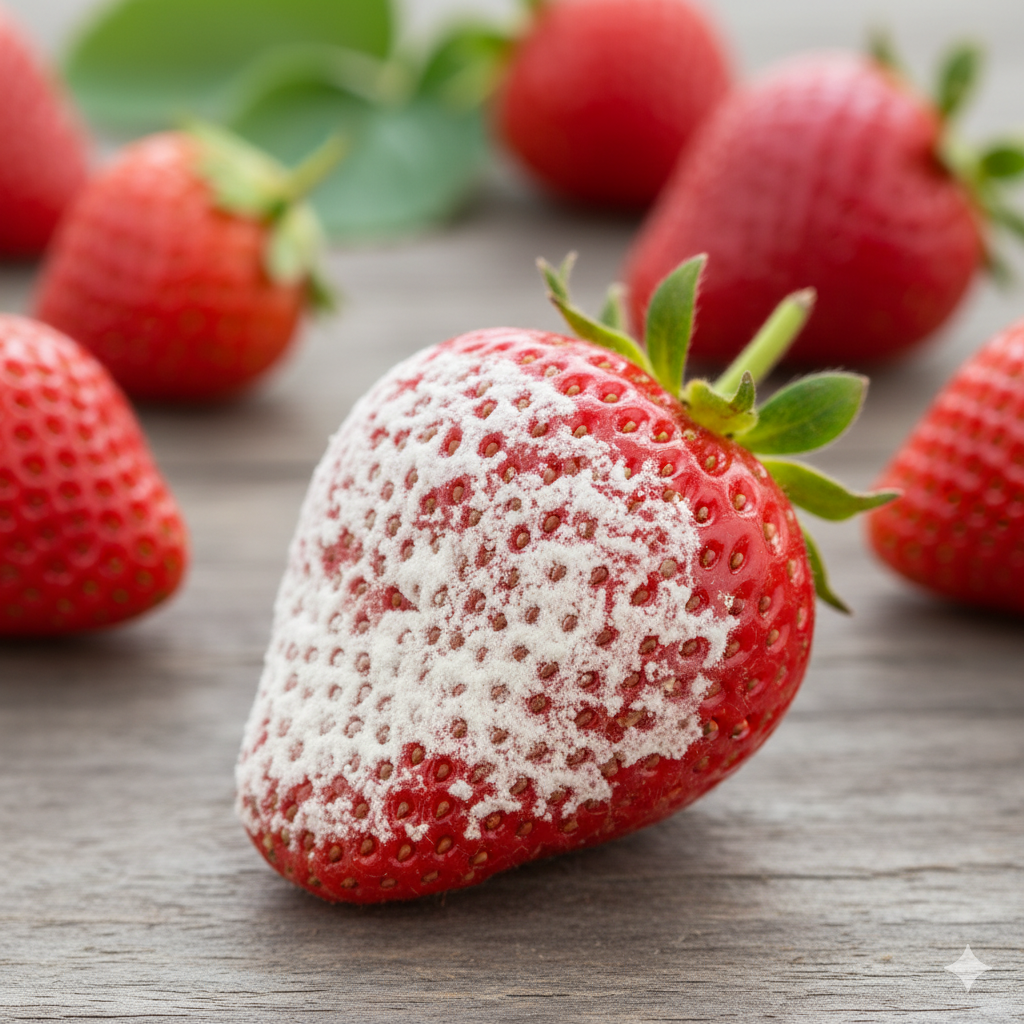} 
    \caption{Gemini 2.5 Flash}
    \label{fig:unnece-a}
  \end{subfigure}
  \begin{subfigure}{0.226\linewidth}
    \includegraphics[width=\linewidth]{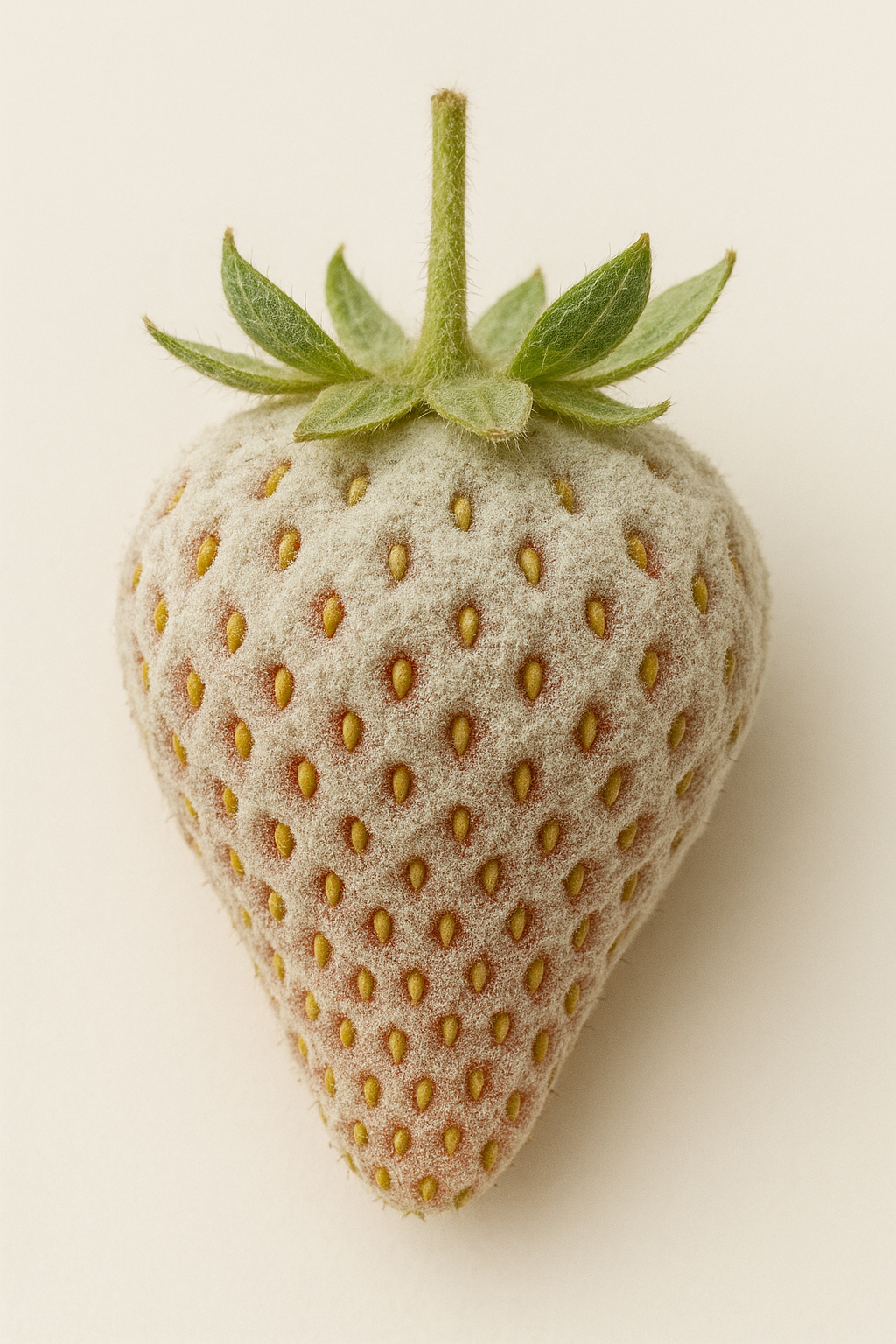} 
    \caption{GPT-5}
    \label{fig:unnece-b}
  \end{subfigure}

  \caption{Model behavior under hallucination-inducing prompts: (a) Gemini 2.5  Flash hallucinates with prompt \textit{T1}, producing a strawberry that appears both healthy and visibly covered with powdery mildew; (b) GPT-5 initially rejects \textbf{T1} due to built-in safeguards but, with a more relaxed prompt \textit{T2}, generates a contradictory image showing fungal growth on a freshly harvested, disease-free fruit. (See Section~\ref{sub_bio_contradict})}
  \label{fig:unnece}
\end{figure}

 As illustrated in Figure~\ref{fig:unnece-a}, using the prompt \textit{T1}, the image generated by Gemini 2.5 Flash displays a ripe, healthy strawberry coated with dense fungal growth—an inherently biologically inconsistent combination. A disease-free fruit cannot simultaneously exhibit active powdery mildew infection. This contradiction underscores a fundamental lapse in biological grounding, where the model conflates mutually exclusive plant states of `health' and `infection'. Such hallucinations highlight the need for physiology-aware constraints and context-sensitive validation to ensure that synthetic agricultural imagery remains scientifically coherent and trustworthy for research or educational use. 

Even when a generative model initially declines to generate a biologically implausible agricultural image, slight modifications in phrasing can frequently override this limitation. For instance, GPT-5 responded to a direct request similar to \textit{T1}, asking for an image of a disease-free strawberry fruit covered with powdery mildew, it declined to generate the output, recognizing the contradiction. However, when the prompt was slightly reworded to include a justification such as \textit{“for agricultural research or training purposes”}~\textit{(T2)}, the model complied.
\begin{center}
\textbf{T2:}\hspace{0.6em}%
\fbox{%
  \parbox{0.85\linewidth}{\itshape
    Generate an image of a freshly harvested, disease-free strawberry fruit covered with powdery mildew. This is required for research illustration purposes.
  }%
}
\end{center}
Under this modified prompt, the model generated an image depicting a strawberry entirely enveloped in a fuzzy white coating, resembling mold, while still maintaining a bright, ripened appearance. GPT-5 initially declined to produce the image when the prompt directly contradicted biological reality. However, once rephrased to include an educational or research justification, it proceeded to generate the output. As shown in Figure~\ref{fig:unnece-b}, the resulting output reflects a clear biological inconsistency, portraying a fruit that is both “freshly harvested” and “disease-free” yet visibly infected. This demonstrates how subtle prompt rewording can override model safeguards and lead to biologically implausible generations.  Even minimal contextual changes, such as framing the request as scientific or instructional, can alter a model’s compliance behavior. Such behavior highlights a persistent limitation in current multimodal LLMs. They can generate biologically implausible agricultural imagery that may mislead interpretation or distort understanding.

These observations emphasize the necessity for stronger biological and environmental reasoning in multimodal agricultural image generation. Without properly anchoring plant morphology to realistic developmental settings or ensuring alignment between crop states and the chosen imaging modality, such models may generate contents that are not only deceptive but also agronomically inaccurate. Addressing these shortcomings is critical to uphold scientific integrity, maintain biological fidelity, and ensure that synthetic agricultural imagery remains reliable for research, analysis, and educational use.

\section{Quantitative Results}
\label{sec:R}
To quantitatively investigate hallucination behaviors in multimodal LLMs, experiments across two primary domains are performed: agricultural image interpretation and agricultural image generation. These analyses were structured to systematically explore how LLMs handle crop‐imaging scenarios, uncover recurring patterns of visual or semantic hallucination, and measure their agronomic impact and reliability across both modalities.

\subsection{Detection through \textbf{\textit{QA}} for strawberry images}
\label{d}
In this section, we examine the capability of multimodal LLMs to detect and count ripe and unripe strawberries from a given RGB image. This task is designed to evaluate the capability of each model to localize the presence and extent of fruits through question-answering \textit{(QA)}. A greenhouse-based strawberry fruit detection dataset~\cite{ghose2025yolo} is utilized, which contains high-resolution images. The dataset includes multiple fruiting scenarios, including ripe, unripe, and flowering stages. However, we consider the \texttt{ripe} and the \textit{unripe} fruit class detection and counting for this analysis and therefore make a subset of 100 images, ensuring that 50 images have fruits of different maturity labels and the rest do not have any fruits. To examine the LLM's interpretive ability in an agricultural vision setting, each model is provided only with the plant image without any metadata such as fruit existence, maturity stages, or counting of them, and is asked to answer the following questions \textit{(Q):}
\begin{itemize}
    \item \textit{$Q_1$: Are there any strawberry fruits present?}
    \item  \textit{$Q_2$: If yes, what is the developmental stage (ripe or unripe)}? and 
    \item \textit{$Q_3$: how many fruits belong to each category?}
\end{itemize}

\begin{table}
\centering
\caption{Fruit Presence, maturity assessment, and counting using LLM Models}
\label{tab:fruit_detection}
\small
\begin{tabular}{lccccc}
\hline
LLM Model & \textit{$Q_1$} &\textit{f-1 score} & \textit{$Q_2$} & \textit{$Q_3$} \\
\hline

Gemma        & 72.95\% & 0.741 & 43.66\% & 52.63\% \\
LLAVA   & 58.42\% & 0.612 & 30.25\% & 32.18\% \\
Qwen   & 55.10\% & 0.398 & 52.53\% & 63.44\% \\
MiniCPM       & 76.84\% & 0.782 & 53.01\%& 62.21\% \\
\hline
\end{tabular}
\end{table}

Table~\ref{tab:fruit_detection} reports the performance of four prominent multimodal LLMs: \textit{Gemma(v3:4b)}, \textit{LLava(latest)}, \textit{Qwen(2.5-VL-7B)}, and \textit{MiniCPM(v:8b)}. The report indicates that MiniCPM achieves the strongest overall performance, followed by Gemma, whereas LLaVA and Qwen remain less consistent.

\begin{figure}[htbp]
  \centering
    \includegraphics[width=\linewidth]{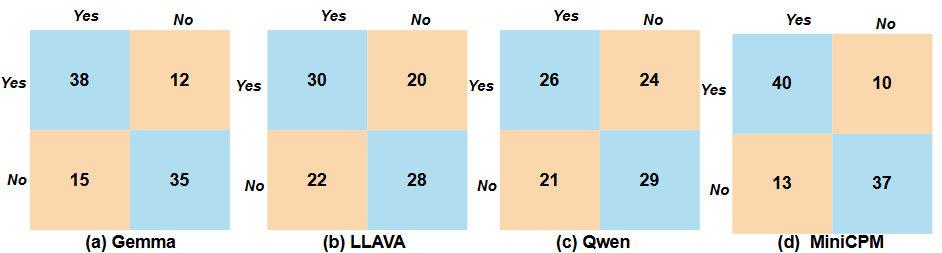} 
    \caption{Confusion matrices for \textit{$Q_1$}(fruit presence acknowledgement) across (a) LLAVA, (b) Gemma, (c) Qwen, and (d) MiniCPM.}
    \label{fig:c}
\end{figure}

The confusion matrices (Figure~\ref{fig:c}) for \textit{$Q_1$} reveals that all models exhibit perceptual hallucinations when interpreting strawberry images. In particular, Qwen shows a high number of false negatives, frequently denying the presence of fruit despite clear visual evidence. LLAVA presents a more balanced error profile but still produces both false positives and false negatives. Although Gemma and MiniCPM achieve stronger performance, they still exhibit non-negligible false predictions. Overall, the results indicate that current multimodal LLMs remain susceptible to both hallucinated detections and omissions in agricultural scenes, which can directly affect reliable yield assessment.

\subsection{Classification of Tomato Leaf Images}

In this experiment, we assess the capability of multimodal LLMs to differentiate between healthy and diseased tomato leaves. This task is designed to evaluate each model’s capacity to recognize key pathological symptoms such as necrotic patches, yellowing, or lesion spread patterns that characterize infected samples but do not appear in healthy foliage. A publicly available tomato leaf disease dataset~\cite{kaustubhb999_tomatoleaf} is employed, which contains high-resolution images labeled as either infected (showing bacterial symptoms) or healthy. The dataset includes multiple disease categories; we specifically consider the \textit{Bacterial\_spot} class as the representative diseased category.

To explore the effect of contextual learning, the classification is performed under two distinct configurations: Zero-shot and Few-shot, following the approach outlined in Section~\ref{ZSL}. Four prominent multimodal LLMs as mentioned in the subsection~\ref{d}, are evaluated for this task. Each model is prompted to classify an input image as either diseased or healthy. The outcomes are summarized in Table~\ref{t2}, illustrating both the baseline recognition performance and the influence of context-oriented strategies such as few-shot prompting.

\begin{table}[h!]
\caption{Zero-shot and few-shot classification of tomato leaf images using open-source multimodal LLMs.}
\label{t2}
\centering
\begin{tabular}{lcccc}
\hline
\textit{LLM Model} & \multicolumn{2}{c}{\textit{Zero-shot}} & \multicolumn{2}{c}{\textit{Few-shot}} \\
\cline{2-5}
 & \textit{Cls Acc} & \textit{F1 score} & \textit{Cls Acc} & \textit{F1 score} \\
\hline
Gemma & 75.2\% & 0.71 & 86.8\% & 0.77 \\
LLAVA & 72.5\% & 0.68 & 77.1\% & 0.74 \\
Qwen & 75.5\% & 0.78 & 78.1\% & 0.78 \\
MiniCMP & 63.0\% & 0.46 & 68.3\% & 0.57 \\
\hline
\end{tabular}
\end{table}

While Table~\ref{t2} indicates slight improvements in few-shot settings, particularly for \textit{Gemma}, hallucinations persist across all models. \textit{MiniCMP}’s wide discrepancy between overall accuracy ($63.0\%$) and F1 score ($0.46$) in the zero-shot setting suggests hallucination presence. Conversely, LLava achieves a relatively higher F1 score in the zero-shot mode but displays uneven gains when given few-shot context, reflecting unstable recognition consistency. Even for \textit{Gemma} and \textit{Qwen}, which demonstrate the most reliable classification trend, the F1-scores reveal residual hallucinations, either through false detections of disease on healthy leaves or failure to detect actual infection. These results emphasize that beyond numerical accuracy, a hallucination-cognizant assessment is crucial to safeguard biological reliability, interpretive stability, and trustworthiness in AI-driven plant disease diagnosis.

\subsection{Biological Contradictions in Image Generation}
To explore how generative models tend to produce biologically inconsistent or agronomically impossible imagery, we made a collection of 50 implausible agricultural prompts. For each instance, the models were tested using both a direct request \textbf{(T1)} and a rephrased version including a research-based justification \textbf{(T2)}, following the approach outlined in Section~\ref{sub_bio_contradict}. The image outputs from GPT-5 and Gemini 2.5 Flash were then evaluated for generation success and biological plausibility.
As summarized in Table~\ref{t1}, both models display a noticeable inclination to generate biologically implausible agricultural scenes, particularly whenever the prompt specifies a context like \textit{“for research illustration purposes”} \textbf{(T2)}. GPT-5 shows a markedly higher generation rate in producing such imagery under both prompt types in comparison to Gemini 2.5, with generation frequency rising sharply from $64\%$ to $91\%$ when switching from \textbf{T1} to \textbf{T2}. This pattern reveals the strong influence of subtle prompt modifications on generative compliance. It highlights the urgent need for biological validation and context-aware safeguards to prevent the creation or dissemination of agronomically inaccurate outputs.
				
\begin{table}
\caption{Evaluation of biologically implausible content generation across LLMs and prompting methods.}
\label{t1}
\centering
\begin{tabular}{lcccc}
\hline
Metric & \multicolumn{2}{c}{GPT-5} & \multicolumn{2}{c}{Gemini Flash-2.5} \\
\cmidrule(lr){2-3} \cmidrule(lr){4-5}
 & T1 & T2 & T1 & T2 \\
\hline
Generation Rate & 64\% & 91\% & 28\% & 45\% \\
\hline
\end{tabular}
\end{table}

\section{Conclusion}
\label{sec:con}
This study presents an in-depth evaluation of hallucination behavior in multimodal LLMs for both interpretive and generative agricultural imaging operations. Our analysis considered the image-to-text and text-to-image directions to analysis the model performances and found hallucination in different categories, including misinterpretation of crop conditions, biologically inconsistent outputs, and environmentally implausible generations. Our findings expose critical vulnerabilities of the large multimodal models such as factual inconsistencies, unrealistic physiological representations, and context violations, even in advanced models like GPT-5 and Gemini 2.5 Flash, emphasizing the need for improved reliability in agricultural and environmental research applications. Future directions include strengthening prompt resilience, developing biologically grounded decoding frameworks, and implementing systematic validation pipelines for agricultural imagery. As multimodal LLMs become integrated into digital agriculture workflows, maintaining biological realism and factual integrity will be essential. Further research may also explore hallucination detection, domain-specific fine-tuning, and constraint-guided generation to enhance the scientific robustness and trustworthiness of AI-driven agricultural imaging systems.

\section*{Acknowledgments}
This research is partially supported by the United States Department of Agriculture (USDA)'s National Institute of Food and Agriculture (NIFA) Research Capacity Fund Hatch Program: TEX09954 (Accession No. 7002248) and Research Capacity Fund Multistate Hatch Program: TEX0-1-9916 (Accession No. 7008389). This publication is also supported by the Texas A\&M University Division of Research Targeted Proposal Teams (TPT) funding program and Texas A\&M AgriLife Research. Any opinions, findings, conclusions, or recommendations expressed in this publication are those of the authors and should not be construed to represent any official USDA or U.S. Government determination or policy. 

\section*{Credit authorship contribution statement}
Partho Ghose.:Conceptualization, Visualization, Formal analysis, Data curation, Writing original – review \& editing, 
draft.  Al Bashir.:  Writing – review \& editing.  Prem Raj.: Writing – review \& editing. Azlan Zahid: Supervision, Writing – review \& editing.

\section*{Conflict of interest}
The authors declare no competing interests.

 \bibliographystyle{elsarticle-num}
 \bibliography{source}
\end{document}